\documentclass[11pt]{article}
\usepackage{geometry}
\usepackage[utf8]{inputenc} 
\usepackage{url}            
\usepackage{booktabs}       
\usepackage{amsfonts}       
\usepackage{nicefrac}       
\usepackage{microtype}      
\usepackage{multirow}
\usepackage{multicol}
\usepackage[dvipsnames]{xcolor}
\usepackage{graphicx}
\usepackage{subcaption}
\usepackage{gensymb}
\usepackage{amsmath}
\usepackage{enumitem} 
\usepackage{amsmath}
\usepackage{amssymb}
\usepackage{mathtools}
\usepackage{float}
\usepackage{parskip}
\usepackage{enumitem}
\usepackage{amsthm}
\usepackage{stix}
\usepackage{hyperref}
\usepackage{natbib}
\usepackage{authblk}

\hypersetup{
    colorlinks=true, 
    linkcolor=Black, 
    urlcolor=Black, 
    citecolor=Black, 
}

\title{Building reliable sim driving agents by scaling self-play}

\author[1]{Daphne Cornelisse\thanks{Corresponding author: cornelisse.daphne@nyu.edu}}
\author[1]{Aarav Pandya}
\author[1]{Kevin Joseph}
\author[2]{Joseph Suárez}
\author[1]{Eugene Vinitsky}
\affil[1]{NYU Tandon School of Engineering}
\affil[2]{Puffer.ai}

\begin{document}

\maketitle

\begin{abstract}
Simulation agents are essential for designing and testing systems that interact with humans, such as autonomous vehicles (AVs). These agents serve various purposes, from benchmarking AV performance to stress-testing system limits, but all applications share one key requirement: reliability. To enable sound experimentation, a simulation agent must behave as intended. It should minimize actions that may lead to undesired outcomes, such as collisions, which can distort the signal-to-noise ratio in analyses. As a foundation for reliable sim agents, we propose scaling self-play to thousands of scenarios on the Waymo Open Motion Dataset under semi-realistic limits on human perception and control. Training from scratch on a single GPU, our agents solve almost the full training set within a day. They generalize to unseen test scenes, achieving a 99.8\% goal completion rate with less than 0.8\% combined collision and off-road incidents across 10,000 held-out scenarios. Beyond in-distribution generalization, our agents show partial robustness to out-of-distribution scenes and can be fine-tuned in minutes to reach near-perfect performance in such cases. We open-source the pre-trained agents and integrate them with a batched multi-agent simulator. Demonstrations of agent behaviors can be viewed at \url{https://sites.google.com/view/reliable-sim-agents}, and we open-source our agents at \url{https://github.com/Emerge-Lab/gpudrive}.
\end{abstract}

\section{Introduction}

Simulation agents are a core part of safely developing and testing systems that interact with humans, such as autonomous vehicles (AVs). In the context of self-driving, these agents, also referred to as road user behavior models, serve two primary purposes: establishing benchmarks for AV behavior \citep{engstrom2024modeling}, and representing other road users in simulators to enable statistical safety testing in both nominal and rare, long-tail scenarios \citep{corso2021survey, montali2024waymo}. While each use case brings particular requirements, \textit{reliability} is an important one that they share.

A reliable simulation agent consistently behaves as intended by the designer, minimizing unintended actions. For instance, agents designed to stress-test AVs should reliably initiate realistic near-collision events, generating safety-critical scenarios to provide meaningful information about the system's behavior in edge cases. Conversely, nominal agents must focus on replicating typical road behavior to simplify experiments that vary other environmental factors, such as weather. In either case, unreliable sim agents introduce \textit{noise} into the evaluation process by producing trajectories that crash too infrequently in the stress-test case and too frequently in the nominal case.

How can we build sim agents that are \textit{close enough}\footnote{Here, close enough is emphasized because what constitutes an acceptable model of human behavior depends highly on the use case.} to reality while maximizing designer specifications, i.e., reliability? One approach relies on generative models, which have shown remarkable progress in producing diverse, human-like behaviors through imitation learning from demonstrations \citep{xu2023bits, DBLP:conf/iclr/PhilionPF24, huang2024versatile}. However, it is uncertain whether they meet the reliability standards of a fully automated AV development pipeline. This is highlighted by the best-performing models in the Waymo Open Sim Agent Challenge \citep[WOSAC]{montali2024waymo}, a well-known benchmark for realistic nominal road user behavior. While state-of-the-art models in the 2024 challenge closely replicate logged human trajectories and achieve high scores on various distributional metrics, they still fall short in critical areas. Ground-truth human trajectories in the dataset rarely or never involve collisions or off-road movements; the best aggregate collision and off-road rates among the top WOSAC 2024 are approximately 4\% and 1-2\%, respectively \citep[CATK, VBD, BehaviorGPT]{zhang2024closed, huang2024versatile, zhou2024behaviorgpt}. This is well below the capabilities of human drivers.

\begin{figure*}[!htbp]  
    \centering
    \includegraphics[width=1\linewidth]{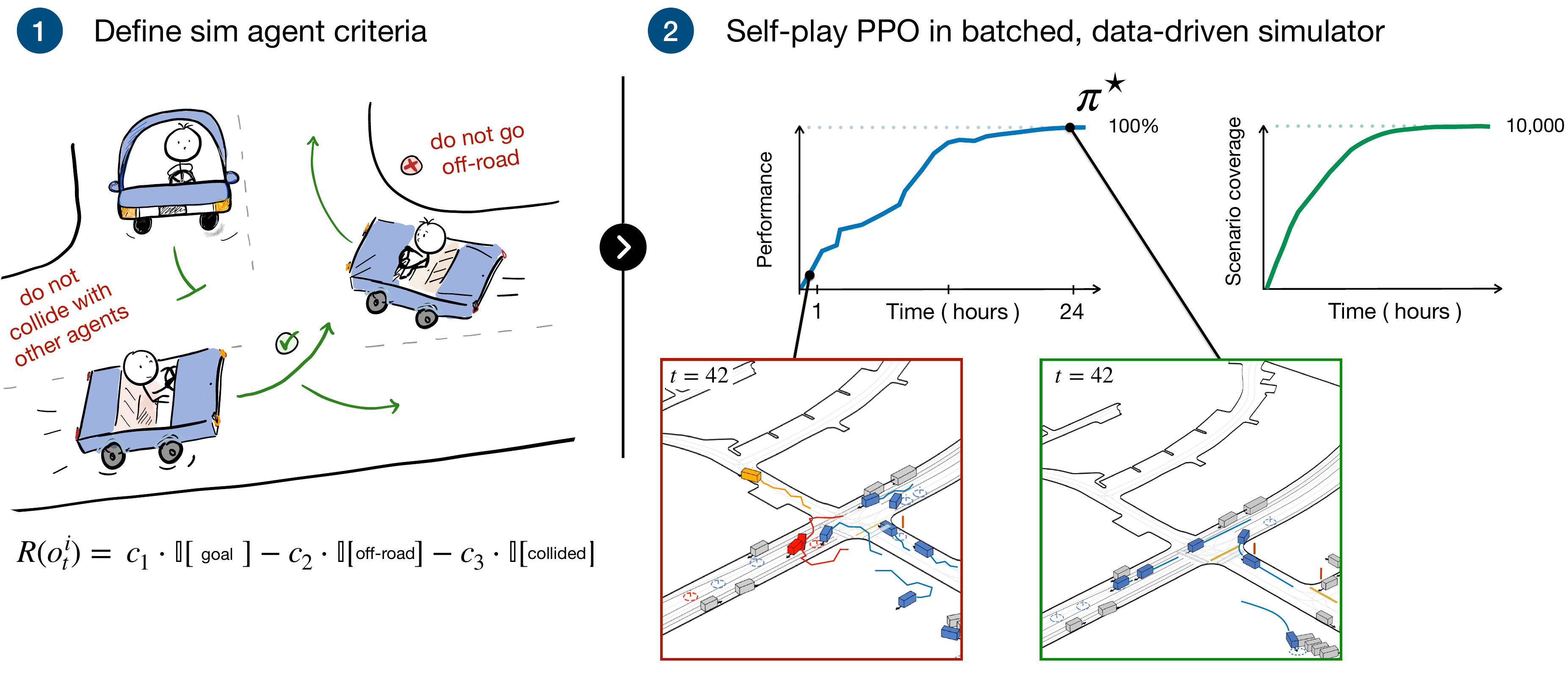}  
    \caption{\textbf{Overview of approach.} \textit{Left}: We define several criteria to guide the learning of simulation agents through rewards. The reward function is a weighted combination of these criteria: $r(o^i_t )= \sum_i c_i \cdot \mathbb{I}[\text{criteria}_i]$. Here, we focus on achieving goal-directed nominal sim agent behavior—ensuring agents stay on the road and avoid collisions while navigating to a target position. \textit{Right}: Over 24 hours on a single GPU, we iterate through 10,000 scenarios (green curve) from the Waymo Open Motion Dataset in GPUDrive \citep{kazemkhani2024gpudrive}, reaching high performance (blue curve, reliability) on the defined criteria after 2 billion agent steps by self-play PPO. The example scenarios illustrate agent behavior at different stages of training. Initially, agents display random behavior and frequently collide with each other and the road edges (marked in orange and red), but their behavior becomes streamlined over many iterations.}
    \label{fig:hero_figure}
\end{figure*}



Importantly, these relatively high failure rates may limit the scalability of AV evaluation and development, especially as generative models are increasingly used to create rare safety-critical scenarios that are underrepresented in real-world data \citep{mahjourian2024unigen}. When trajectories deviate unpredictably, researchers or engineers must find out: Is the observed outcome a signal or an artifact of simulator noise? For example, if 1 in 10 scenarios reflects unintended behavior, distinguishing meaningful failures from artifacts becomes a time-consuming task. As such, making sim agents more reliable seems to be a key pillar to further scale AV evaluation and development.

The question becomes: How can we close this reliability gap for state-of-the-art sim agents? Assuming we can precisely define what the agent should adhere to (e.g., stay on the road), there is reason to believe that self-play reinforcement learning (RL) is a piece of the puzzle. Evidence from a broad body of recent literature on games shows that self-play RL, combined with well-defined criteria (e.g., maximize score X) can produce agents capable of perfect, superhuman gameplay in the large compute and data regime \citep{silver2018general, DBLP:conf/iclr/Bakhtin0LGJFMB23, openai5}.

We study whether self-play at scale improves the reliability of sim agents. Specifically, we ask:
\begin{enumerate}[noitemsep]
    \item How does the \textit{reliability} (as measured by performance on the test set of metric X) of sim agents through self-play scale as a function of the data available?
    \item How well do these agents generalize to unseen scenarios and out-of-distribution events?
\end{enumerate}
To explore these questions, we train agents via self-play using a semi-realistic human perception framework in a data-driven simulator \citep{kazemkhani2024gpudrive}. We evaluate performance across thousands of scenarios from the Waymo Open Motion Dataset \citep{ettinger2021large}. We find that self-play PPO scales effectively with on-policy data and compute. After sufficient training, models generalize well to 10,000 unseen test traffic scenarios, effectively closing the train-test gap.

At scale, self-play PPO sim agents consistently achieve the specified criteria (Section \ref{sec:task_definition}): staying on the road, avoiding collisions, and reaching a target position. This establishes a framework where agents can be tuned to achieve specific collision rates, enabling both nominal and safety-critical traffic simulation. By improving the reliability standards of sim agents, our approach supports the continued scaling and automation of AV development and evaluation pipelines.

Finally, we take a step towards fine-tuning these agents for behaviors underrepresented in the dataset, a useful capability for safety-critical applications. To facilitate further research, we open-source the pre-trained agents at \url{www.github.com/Emerge-Lab/gpudrive}, 
allowing others to reproduce our results and seamlessly use these sim agents.

\section{Method}


\subsection{Dataset and simulator}
We conduct our experiments in GPUDrive, a data-driven, multi-agent, GPU-accelerated simulator \citep{kazemkhani2024gpudrive}. GPUDrive contains $K =$ 160,147 real-world traffic scenarios from the Waymo Open Motion Dataset \citep[WOMD]{ettinger2021large}. Each scenario $k \in K$ comprises a static road graph, $R_k$, and a time series of \textit{joint} logged human trajectories: $\mathcal{S}_k = \{(\mathbf{s}_t, \mathbf{A}_t)_{t=0}^{T=90}, R_k \}$ where $\mathbf{s}_t \in \mathbb{R}^{(1, F)}$ represents the world state represented as $F$ features at time $t$, and $\mathbf{A}_t \in \mathbb{R}^{(N, 2)}$ represents the action matrix for all $N$ agents in the scene. The joint agent demonstrations are 9 seconds long and discretized at 10 Hz. 

\begin{figure}[htb]
    \centering
    \includegraphics[width=1\linewidth]{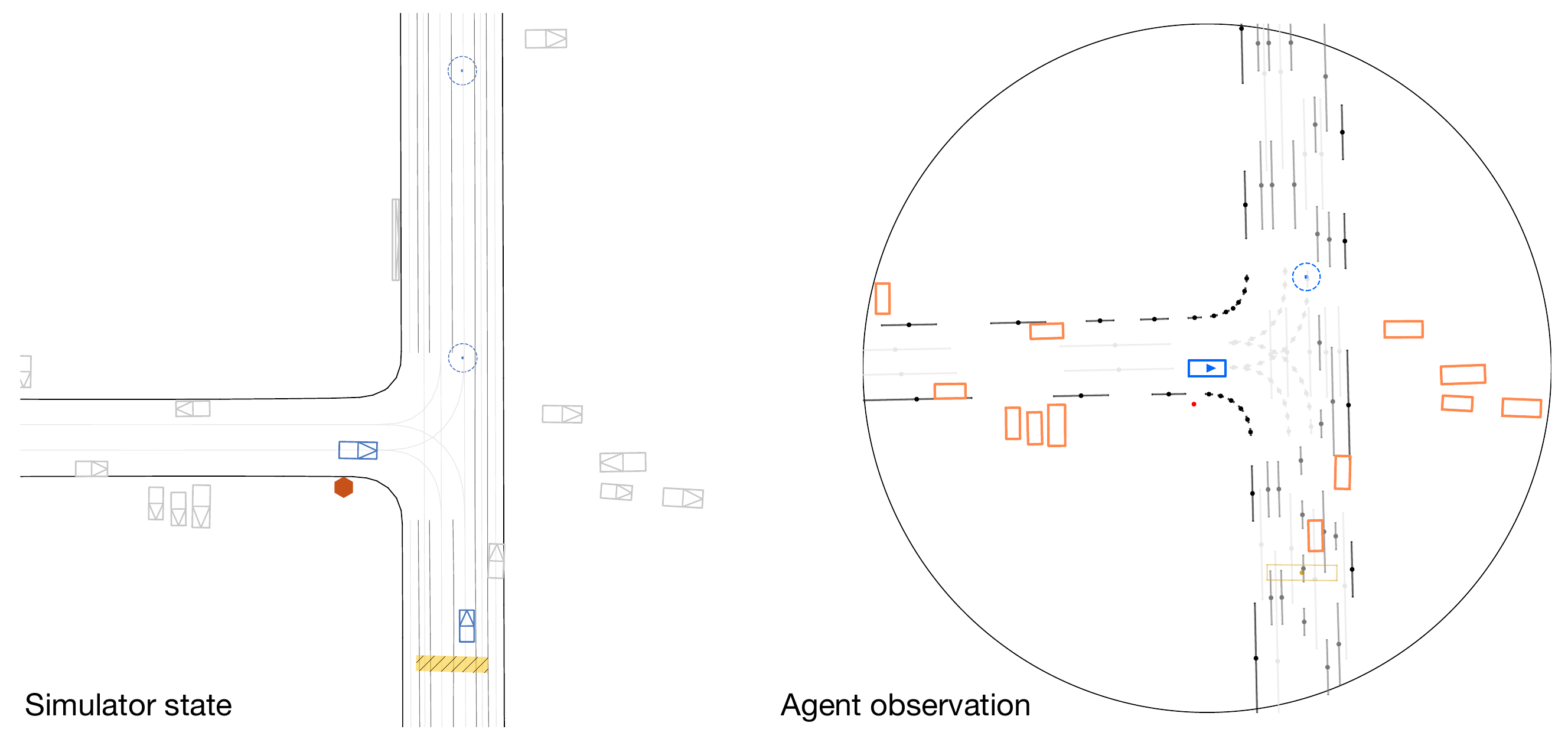}
    \caption{\textbf{Sample scenario state with corresponding agent observation}. \textit{Left}: Example scenario from the Waymo Open Motion Dataset rendered in GPUDrive as shown from a bird's eye view. The boxes ($\textcolor{RoyalBlue}{\hrectangle}$) indicate controlled agents and the circles ($\textcolor{RoyalBlue}{\odot}$) indicate the goal positions for every controlled agent. \textit{Right}: Scene view from the agent in the center ($\textcolor{Periwinkle}{\hrectangle}$). Agents see a subset of the road points within a configurable radius (here $r_o = 50$ meters) and their corresponding types and segment length. Road types are road edges ($\textcolor{Black}{\bullet}$) and road lanes ($\textcolor{Gray}{\bullet}$) They can also view the relative position and velocity of the other agents in the scene ($\textcolor{YellowOrange}{\hrectangle}$). Agents in gray are static throughout the episode as they are parked cars but this information is not visible to the agent i.e. the agent does not know that the gray cars are guaranteed not to move and consequently all cars are orange in the agent observation view.}
    \label{fig:state_and_obs_space}
\end{figure}


\subsection{Task definition and measuring performance}
\label{sec:task_definition}

\subsubsection{Task definition}
We aim to study how the reliability of simulation agents trained via self-play scales with data. To this end, we design a task with well-defined metrics such that experimental results are easy to interpret. Given a traffic scenario \(\mathcal{S}_k\) with \(N\) controlled agents we task every agent to navigate to a designated goal position while satisfying two criteria: (1) avoiding collisions with other agents and (2) staying on the road.  

To obtain valid goals, we use the endpoints \((x^i_T, y^i_T)\) (marked by \(\textcolor{RoyalBlue}{\odot}\) in Figure \ref{fig:state_and_obs_space}) from the WOMD. Agents are initialized from the starting positions \((x^i_0, y^i_0)\) of the WOMD. Given how the WOMD is collected and processed, we know that the human agents in the dataset must have successfully reached their endpoints within 9 seconds (or 91 steps). We assume that in principle, all agents should be capable of doing the same. To reflect this, a scenario is considered solved when \textit{all controlled agents} reach their target positions within 91 steps while adhering to the specified criteria. 

\subsubsection{Metrics}
\label{sec:metrics}

We use four \textit{scene-based metrics} to quantify performance:
\begin{itemize}[noitemsep]
    \item \underline{Goal achieved} \textcolor{RoyalBlue}{↑}: Percentage of agents that reached their target position within $T=91$ steps.
    \item \underline{Collided} \textcolor{BrickRed}{↓}: Percentage per scenario indicating objects that collided, at any point in time, with any other object, i.e. when the agent bounding boxes touch.
    \item \underline{Off-road}  \textcolor{RedOrange}{↓}: Percentage of agents per scenario that went off the road or touched a road edge, at any point in time.
    \item \underline{Other} \textcolor{lightgray}{↓}: Percentage of agents per scenario that did not collide or go off-road but also did not reach the goal position.
\end{itemize}
The \underline{Collided} and \underline{Off-road} metrics align with the Waymo Open Sim Agent Challenge and Waymax \citep{montali2024waymo, gulino2024waymax}. Specifically, \underline{Collided} is part of the ``object interaction metrics" category and the off-road events are part of the ``map-based metrics'' category. Under the assumption that human road users have near zero collision and off-road events, we can meaningfully compare our scores to the top submissions \citep{huang2024versatile, zhou2024behaviorgpt} \footnote{Technically, WOSAC frames this as a distribution-matching problem: metrics are first computed as event counts, which are then compared to the distribution of log replay trajectories across several rollouts.}. 

The \underline{Goal achieved} metric is not directly reported in WOSAC, making it less comparable. The most similar metric is the Route Progress Ratio used in Waymax \citep{gulino2024waymax}, which measures how far an agent travels along the logged trajectory. However, since our focus is not on mimicking logged trajectories but on precisely reaching a particular goal, a binary metric is, in our case, a more meaningful indicator of performance. However, reaching the goal roughly corresponds to a Route Progress Ratio of $100\%$.

\textit{Agent-based metrics}: Since the scene-based metrics are biased towards scenes with a small number of agents (one agent colliding in a scene with 2 agents vs. 10 scenes provides a fraction of 1/2 vs 1/10th), we also report the metrics above in \textit{agent-based} way, where we aggregate the counts across the whole dataset and then divide them by the number of total agents. 

In both cases, the ceiling for this task is 100\% \underline{Goal achieved}, 0\% \underline{Collided}, and 0\% \underline{Off-road}. 

\subsection{State and observation space}

This section outlines the design choices and parameterization of the observation \( \mathbf{o}_{t}^i \) for agent \( i \) at time \( t \). We make these choices to reflect semi-realistic limits on human perception. The observation encodes the agent’s partial view of the scenario state \( s_t \), capturing the information necessary for decision-making. In this work, we model the RL problem as a Partially Observed Stochastic Game \citep[][POSG]{posg04}, where agents make simultaneous decisions under partial observability. We briefly describe the core aspects of the observation space and give a complete overview in Appendix \ref{appendix:feature_details}. Design choices:
\begin{itemize}
    \item \textbf{Relative coordinate frame:} All agent information is presented in an ego-centric coordinate frame to align with human-like perception.
    \item \textbf{Observation radius:} The observation radius \( r_o \) determines the visible area around the agent. For our experiments, we set \( r_o = 50 \) meters, as illustrated in Figure \ref{fig:state_and_obs_space}.
    \item \textbf{No history:} Agents only receive information from the current timestep.
    \item \textbf{Road graph:} We reduce the full road graph, which consists of up to 10,000 sparsely distributed road points, for computational efficiency. To reduce the number of points corresponding to straight lines, we run the polyline reduction threshold of the polyline decimation algorithm \citep{visvalingam2017line} in GPUDrive to 0.1 which roughly cuts the number of points by a factor of 10. We also cap the maximum visible road points at 200, selecting 200 points from those in the view radius in a random order if there are more than $200$ points, creating a sparse view of the local road graph. Empirical results show this is sufficient for agents to navigate toward goals without going off the road or causing collisions.
    \item \textbf{Normalization:} Features are normalized to be between -1 and 1 by the minimum and maximum value in their respective category. Details are found in Tables \ref{tab:ego_feature_dimensions}, \ref{tab:rg_feature_dimensions}, and \ref{tab:partner_feature_dimensions}.

\end{itemize}

\subsection{Action space and dynamics model}

To align with the control outputs of real human road users more closely, we take the action for every agent $i$ to be a vector of the following discrete random variables: $\mathbf{a}^i_t = (\tilde{a}, \tilde{s})$ where acceleration actions are 7 actions defined over an evenly spaced grid between $[-4, 4 ]$ and the steering wheel angle are 13 actions defined over an evenly spaced grid between $[- \pi, \pi]$. The bounds are set to reflect the kinematic constraints of real driving. We assume that the random variables $\tilde{a}, \tilde{s}$ are not independent (e.g. sharp turns are less likely at high acceleration) and model the conditional joint probability mass function (pmf) of the two discrete random variables, where we condition on the current observation of agent $i$ at time step $t$:
$\pi_{\tilde{a}, \tilde{s}}(a, s \mid \mathbf{o}_t^i) := P(\tilde{a} = a, \tilde{s} = s\mid \mathbf{o}_t^i)$. The conditional pmf $\pi_{\theta}$ describes the behavior under the assumption that $\mathbf{o}_t^i$ takes a fixed set of values. The total joint action space contains $7 \times 13 = 91$ actions. With these actions, agents are stepped in the simulator using an Ackermann bicycle model \citep{rajamani2011vehicle}.

\subsection{Reward function}

We define the individual agent rewards as follows:  
\begin{equation*}  
r(\mathbf{o}^i_t, \mathbf{a}^i_t) = w_{\text{Goal achieved}} \cdot \mathbb{I}[\text{Goal achieved}] - w_{\text{Collided}} \cdot \mathbb{I}[\text{Collided}] - w_{\text{Offroad}} \cdot \mathbb{I}[\text{Offroad}]    
\end{equation*}  


\subsection{Collision behavior}

During training and testing, we allow agents to continue the episode even after going off-road or colliding with another agent in the scene. Agents receive a penalty for each collision or off-road event, allowing them to accrue multiple penalties throughout an episode. See Appendix \ref{appendix:collision_behavior} for the details.

\subsection{Model}

We use a neural network with an encoder and a shared embedding, as illustrated in Figure \ref{fig:network_architecture} (see Appendix \ref{sec:neural_net}). The flat observation vector is first decomposed into three modalities: the dense ego state, the sparse road graph, and the sparse partner observations. Each modality is processed independently. Inspired by the late fusion approach in Wayformer \citep{DBLP:conf/icra/NayakantiAZGRS23}, we then concatenate the outputs, apply max pooling, and pass the result through a shared embedding. This hidden embedding is fed into separate actor and critic heads, each implemented as a single feedforward layer. The model only has $\approx 50,000$ trainable parameters.

\subsection{Training}

\paragraph{Self-play PPO} In each scenario, we control up to $N=64$ agents using a shared, decentralized policy $\pi_{\theta}$. Actions are independently sampled from the policy based on the ego views of each agent $i$ during every step in the rollout: $\mathbf{a}_t^i \sim \pi_{\theta}(\cdot \mid \mathbf{o}_t^i)$. We train agents using Proximal Policy Optimization \citep[][PPO]{ppoSchulman17} using batches of $S = 800$ distinct scenarios, with the set of training scenarios uniformly resampled every 2 million steps. Initially, agents exhibit random behavior and crash frequently. Over time, the agents' behavior becomes more streamlined, creating smooth trajectories with high rates of reaching the goals. 


\section{Related work}

\textbf{Self-play for agents in games:}
Self-play RL \citep{5392560, tesauro1995temporal} has been a core ingredient in creating effective agents across a wide range of complex games. Notable examples include superhuman gameplay in two-player zero-sum games like Chess and Go \citep{silver2018general}, expert human-level play in Stratego \citep{perolat2022mastering} and Starcraft \citep{starcraft}, as well many-player games that require some level of cooperation like Diplomacy \citep{DBLP:conf/iclr/Bakhtin0LGJFMB23} and Gran Turismo~\citep{gtsophy}. These successes have demonstrated the effectiveness of self-play, particularly in the large-data, large-compute regime. However, the majority of its successes are in variants of zero-sum games whereas driving tasks are likely general-sum and feature many-agent interaction.

\textbf{RL for driving agents:}
Reinforcement learning has been explored for the design of autonomous driving agents, though state-of-the-art agents are currently far below the human rate of between $800000$ km per police-reported traffic crash in the United States~\citep{stewart2023overview} or as much as $1$ crash per $24800$ km in more challenging domains such as San Francisco~\citep{flannagan2023establishing}. These agents are frequently trained in simulators built atop large open-source driving datasets~\citep{gulino2024waymax,nocturne,kazemkhani2024gpudrive} such as Waymo Open Motion \citep[WOMD]{ettinger2021large}, \citep[NuScenes]{caesar2020nuscenes}, \citep[ONE-Drive]{onedrive} though there are also procedurally generated~\citep{li2022metadrive} and non-data-driven simulators~\citep{carla17}. These datasets collectively add up to tens of thousands of hours of available data and are often used to train RL agents in \emph{log-replay} mode, a setting in which only one agent is learning and the remainder are either replaying human trajectories or executed hand-coded policies. 

The complexity of scaling RL in these settings has led to the creation of batched simulators \citep[GPUDrive]{kazemkhani2024gpudrive}, \citep[Waymax]{gulino2024waymax}, \citep[Gigaflow]{gigaflow} whose high throughput helps ameliorate issues of sample complexity. Many works have explored ways to use these simulators to learn high-quality reinforcement learning agents through RL including uses of self-play \citep{copo,nocturne,closed_loop_driving,closed_loop_v2,aspDrive,jaeger2025carl}. Our work is mostly distinct from these by the scale of training and a significantly lower crash and off-road rate than has previously been observed with the exception of \citep[Gigaflow]{gigaflow}. \emph{Gigaflow} achieves a scale and robustness above what is in this paper, crashing between every hundred thousand to million kilometers in self-play, whereas we crash on the order of every 10 kilometers. However, our work provides complementary insights to \emph{Gigaflow} in several ways 1) the results are open-source, creating models and training scripts that can be reused in other works; 2) our training setup draws scenes from log-replay data, creating a curriculum of both easy and challenging tasks as opposed to scattering goals over a large map as is done in \emph{Gigaflow}.



\section{Results}

\begin{table*}[htbp]
\centering
\caption{Aggregate scene-based performance in \% across $N=10,000$  randomly sampled train and test traffic scenarios from the Waymo Open Motion Dataset (mean $\pm$ std). Metrics are defined in section \ref{sec:metrics}.}
\begingroup
\setlength{\arrayrulewidth}{1pt} 
\renewcommand{\toprule}{\specialrule{0.2pt}{0pt}{0pt}} 
\renewcommand{\midrule}{\specialrule{0.2pt}{0pt}{0pt}} 
\renewcommand{\bottomrule}{\specialrule{0.2pt}{0pt}{0pt}} 
\fontsize{2}{3.5}\selectfont 
\resizebox{\linewidth}{!}{%
\begin{tabular}{@{}lcccc@{}} 
\toprule
\textbf{Dataset} & \textbf{Goal achieved} \textcolor{RoyalBlue}{↑} & \textbf{Collided} \textcolor{BrickRed}{↓} & \textbf{Off-road} \textcolor{RedOrange}{↓} & \textbf{Other} \textcolor{lightgray}{↓} \\
\midrule
Train  & 99.84 ± 1.27 & 0.38 ± 2.91 & 0.26 ± 2.17 &  0.13 ± 1.14 \\ 
Test   &  99.81 ± 1.53 & 0.44 ± 3.17 & 0.31 ± 2.59 & 0.14 ± 1.16 \\ 
\bottomrule
\end{tabular}%
}
\endgroup
\label{tab:aggregate_perf_best_policy}
\end{table*}

\subsection{Scaling with data}
\label{sec:scaling_with_data}

\begin{figure*}[!htbp]  
    \centering
    \includegraphics[width=\linewidth]{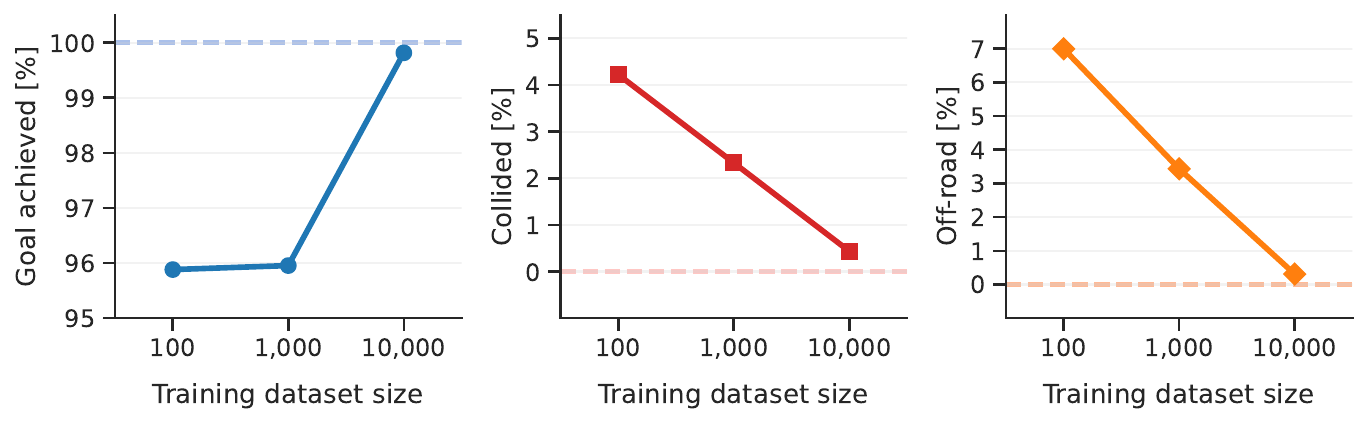}  
    \caption{\textbf{Scaling with data}. Average performance with standard errors on 10,000 unseen scenarios from the WOMD validation set as a function of the training dataset size. The striped lines indicate optimal performance.}
    \label{fig:scaling_laws}
\end{figure*}

\begin{figure*}[!htbp]
\centering \includegraphics[width=1\linewidth]{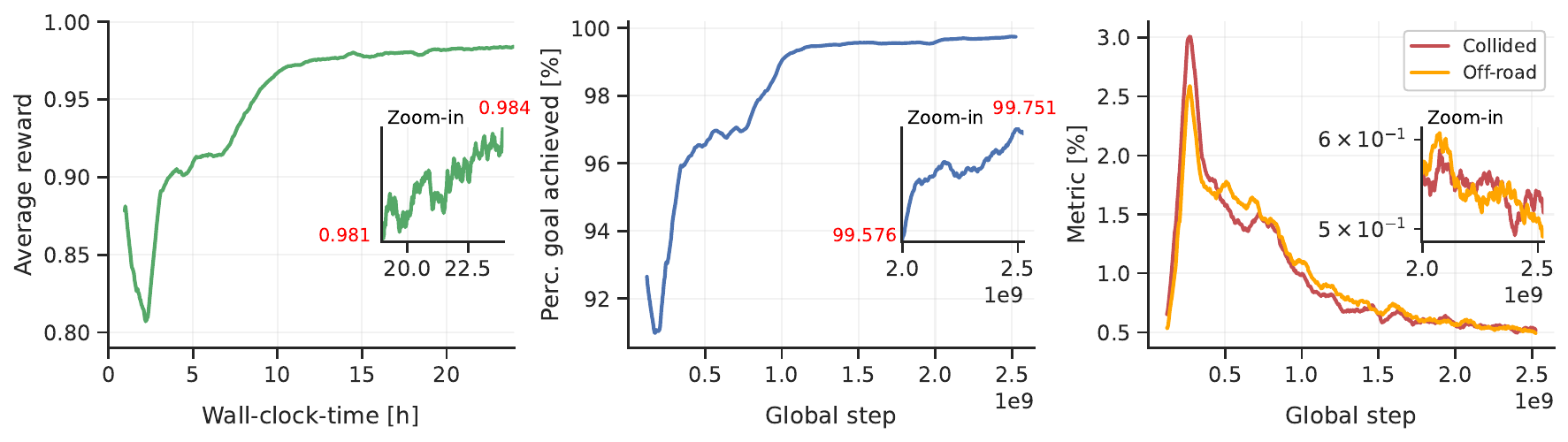}
\caption{\textbf{Batch performance throughout training.} \textit{Left}: Average reward per agent (maximum of 1) as a function of wall-clock time. We train agents for at most 24 hours. \textit{Center}: Goal achievement rate per batch as a function of global steps (2 billion steps generated in 24 hours). \textit{Right}: Percentage of agents that collide with another agent (red) or with a road edge (orange). All curves are smoothed using a rolling window of 250 steps. The inset figures show a zoomed-in view of the final four hours of the run, with the y-axes displayed on a logarithmic scale. The red annotations on the insets indicate the minimum and maximum values within the zoomed-in window. Note that the metrics reported during training are by excluding trivial agents, we only control agents that have to drive for more than 2 meters to reach their goal destination.} 
\label{fig:train_performance_log} 
\end{figure*}

\paragraph{Solving the full Waymo Open Motion Dataset under partial observability}
We investigate whether agents with a partial view of the environment can solve all scenarios in the Waymo Open Motion Dataset. Our results show that almost all scenarios can be solved successfully. After $2$ billion training steps, agents achieve a goal-reaching rate of $99.84$\%, a collision rate of $0.38$\%, and an off-road rate of $0.26$\% on the training dataset.\footnote{These metrics are computed in alignment with the way they are defined in WOSAC, but it should be noted that this is an over-optimistic metric as it includes many agents that simply need to remain in place as they are initialized right next to their goals. Excluding such agents and only controlling agents that must drive more than 2 meters before reaching their goal gives performance metrics of: $99.40$\% goal-reaching rate, $0.5$\% collision rate, and $0.6$\% offroad rate.} Furthermore, as depicted in Figure \ref{fig:train_performance_log}, zooming in on the final four hours of training suggests that metrics exhibit a continued, albeit gradual, improvement, indicating that performance can be further increased with additional training. This training run took $24$ hours on a single NVIDIA A100 GPU.


The agent-based metrics are similar to the scene-based metrics reported above: a goal rate of $99.72$\%, a collision rate of $0.26$\%, and an off-road rate of $0.35\%$. Sample rollouts with the best-trained policy are shown in Figures \ref{fig:3d_examples}, \ref{fig:2d_examples} and on the project page.

\paragraph{Effective generalization to unseen scenarios with sufficient data}
We conduct experiments with 100, 1,000, 10,000, and 100,000 unique training scenarios to assess how self-play performance scales with the diversity of training scenes. Table \ref{tab:aggregate_perf_best_policy} summarizes the results. We find no significant train-test gap when training with 10,000 scenarios or more, indicating the model generalizes well to new, unseen situations. Figure \ref{fig:scaling_laws} shows the key metrics as a function of training dataset size. Notably, with 10,000 training scenarios, the model reaches nearly the ceiling of our benchmark, achieving a 99.81\% goal-reaching rate, 0.44\% collision rate, and 0.31\% off-road rate on 10,000 held-out test scenarios. Note that these numbers do not add to 1 as a vehicle may collide and go off-road in the same step.

\subsection{Extrapolative generalization and fast fine-tuning}
\label{sec:finetuning}

\begin{figure*}[htbp]  
    \centering
    \includegraphics[width=\linewidth]{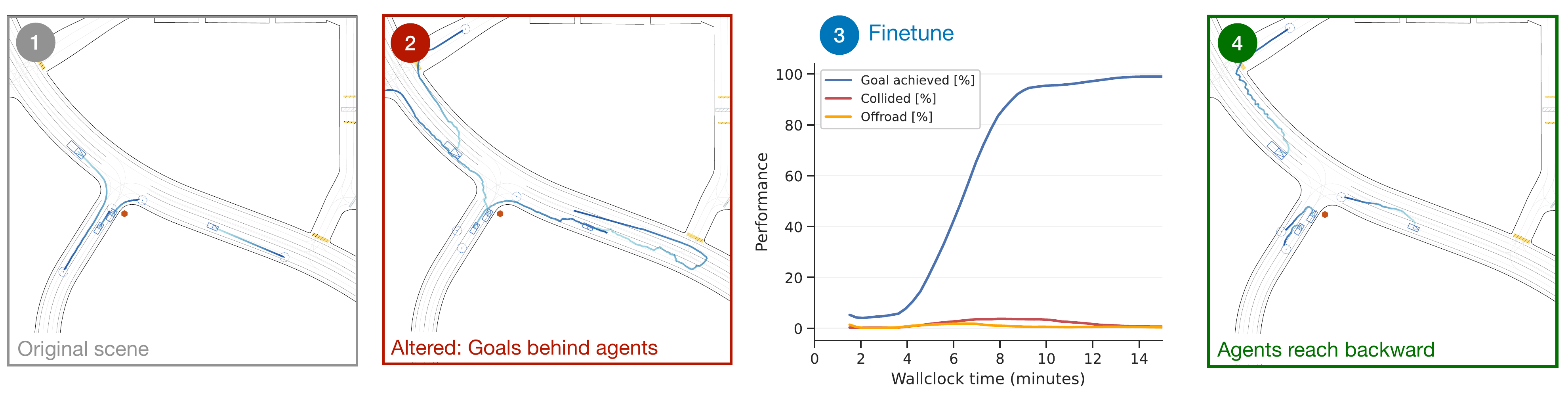}  
    \caption{\textbf{Fine-tuning agent behaviors} \underline{1:} In most scenarios, agent target positions are located in front of them. The figure shows a typical example from the dataset with rollouts from the trained policy. \underline{2:} Fewer than 2\% of agent goals require backward driving or a U-turn. To evaluate agent performance in such out-of-distribution cases, we create hand-designed scenarios where goals are placed behind agents. As expected, performance drops significantly (by 50\%), as agents struggle to reach these goals. In this scene, only a single agent achieves its new goal. \underline{3:} To address this, we fine-tune a model pre-trained on 10,000 WOMD scenarios using the 13 hand-designed cases. Within 15 minutes, agents successfully learn to navigate to the goals behind them. \underline{4:} A rollout of the fine-tuned model demonstrates its ability to handle the altered scenario. Each agent executes a U-turn to get to its goal.}
    \label{fig:finetuning_steps}
\end{figure*}

\paragraph{Navigating backwards} 
Beyond generalization to within distribution scenarios, as reported in Section \ref{sec:scaling_with_data}, we are interested in agent performance in out-of-distribution events. This is useful to know, as researchers may typically manipulate scenarios or make them harder in some way to test the limits of AV systems. Where do these agents break, and how easily can they be finetuned? We analyzed the complete training data set to find rare behaviors and found that agents rarely have to reach a goal that is behind them ($1.86\%$, Details in Appendix~\ref{appendix:ood}). We then hand-designed 13 scenarios from the test dataset with a total of 27 agents across all scenes, placing goals behind agents. Figure \ref{fig:finetuning_steps}.2 illustrates an example of such a scene. We summarize the results in Table \ref{tab:goals_behind_agents_perf}. Overall, we observe that the performance drops from 100\% to 53.5\% goal-reaching rate when placing the targets behind the agents. Unsurprisingly, agents exhibit poor performance on events that are rare or entirely unobserved in the training scenarios.


\paragraph{Fast finetuning}
As a proof of concept, we demonstrate how self-play reinforcement learning enables rapid fine-tuning of a model to learn new behaviors, such as navigating backward, using only a few samples. Figure \ref{fig:finetuning_steps} provides an overview of our approach. Initially, introducing a rare and possibly out-of-distribution scenario—where goals are positioned behind agents—leads to a drop in performance ($1 \rightarrow 2$). To address this, we take the 13 hand-designed scenarios and fine-tune the policy that was pre-trained on 10,000 WOMD scenarios ($3$). The model starts with a low goal-reaching rate but quickly adapts, achieving 100\% success within 15 minutes of training. After fine-tuning, agents can reliably reach goals behind them ($4$). We share an accompanying video of before and after finetuning at \underline{\href{https://sites.google.com/view/reliable-sim-agents/home}{the project page}}.

\paragraph{Realism assessment}
\label{sec:realism_wosac}
While our objective in this work is not to develop a \textit{human-like agent}, it is informative to assess how close the resulting agent behavior matches the distributional properties of human driving. To obtain an indication of the agent realism, we evaluate on the Waymo Open Sim Agents Challenge (WOSAC) \citep{DBLP:conf/nips/MontaliLMKRLGEY23}, an established benchmark for assessing realism in driving. In a slight deviation from the standard benchmark, we condition the agents on the \textit{end points} of the log trajectories, instead of conditioning on the first second of logs.

\begin{table*}[htbp]
\centering
\caption{Comparison of model performance across WOSAC evaluation metrics. Arrows indicate whether higher (↑) or lower (↓) values are better. Note that collision/offroad here is a likelihood, not a measure of collision/offroad rate.}

\begingroup
\setlength{\arrayrulewidth}{0.8pt}
\renewcommand{\toprule}{\specialrule{0.2pt}{0pt}{0pt}}
\renewcommand{\midrule}{\specialrule{0.2pt}{0pt}{0pt}}
\renewcommand{\bottomrule}{\specialrule{0.2pt}{0pt}{0pt}}
\resizebox{\textwidth}{!}{%
\begin{tabular}{@{}l|c|c|cccc|ccc|cc@{}} 
\toprule
\multirow{2}{*}{\textbf{Model}} & \textbf{Realism} \textcolor{RoyalBlue}{↑} & \textbf{minADE} \textcolor{BrickRed}{↓} & \multicolumn{4}{c|}{\textbf{Kinematics} \textcolor{RoyalBlue}{↑}} & \multicolumn{3}{c|}{\textbf{Interaction} \textcolor{RoyalBlue}{↑}} & \multicolumn{2}{c}{\textbf{Map-Based} \textcolor{RoyalBlue}{↑}} \\
\cmidrule(lr){2-2} \cmidrule(lr){3-3} \cmidrule(lr){4-7} \cmidrule(lr){8-10} \cmidrule(lr){11-12}
 & Meta & [m] & Lin Speed & Lin Accel & Ang Speed & Ang Accel & Dist Obj & Collis & TTC & Dist edge & Offroad \\
\midrule
Logged Oracle & - & - & 0.476 & 0.478 & 0.578 & 0.694 & 0.476 & 1.000 & 0.883 & 0.715 & 1.000 \\
\textbf{Self-play PPO} & \textbf{0.629} & \textbf{8.205} & \textbf{0.170} & \textbf{0.272} & \textbf{0.302} & \textbf{0.471} & \textbf{0.073} & \textbf{0.970} & \textbf{0.882} & \textbf{0.262} & \textbf{0.910} \\
VBD & 0.720 & 1.474 & 0.359 & 0.366 & 0.420 & 0.522 & 0.368 & 0.934 & 0.815 & 0.651 & 0.879 \\
\bottomrule
\end{tabular}%
}
\endgroup
\label{tab:wosac_comparison}
\end{table*}

Table \ref{tab:wosac_comparison} shows the aggregate scores across 100 held-out scenes. Despite not intentionally optimizing for realism, our method gets a realism meta-score of $0.629$. The simulated collision rate of the agent in WOSAC is 0.5\%, which is comparable to what we report in Table \ref{tab:aggregate_perf_best_policy}. As ranked by the realism meta-metric, it is above a linear extrapolation policy ($0.324$) \citep{montali2024waymo}, and competitive with entries on the 2024 WOSAC challenge, such as VPD-BP ($0.63$), and GigaFlow ($0.6190$) \citep{gigaflow}. 









\section{Discussion and conclusions}

Our results lead us to three main conclusions:
\begin{itemize}
\item \textbf{Self-play at scale reliably achieves well-defined criteria in unseen scenarios:}  
Our findings suggest that self-play RL scales effectively with available data (Section \ref{sec:scaling_with_data}), achieving low-collision performance on the Waymo Open Motion Dataset (WOMD) with no train-test gap. To the best of our knowledge, this is the first demonstration of this level of performance on WOMD. Compared to state-of-the-art supervised models, such as CATK \citep{zhang2024closed}, VBD \citep{huang2024versatile}, and BehaviorGPT \citep{zhou2024behaviorgpt}, our approach appears to reduce the collision and off-road rates by approximately 4 $\times$. Our agents crash on the order of once every 30 minutes, which, while well below human capabilities, represents a meaningful increase over available baselines.
\item \textbf{Rare events remain a challenge.}  
Agents struggle with rare or out-of-distribution scenarios, such as goals placed behind them (Section \ref{sec:failure_modes}) or navigating roundabouts. In these cases, performance drops significantly, indicating that performance on uncommon situations remains a key limitation.  
\item \textbf{Fine-tuning quickly improves performance in unseen scenarios:}  
Fine-tuning on a small subset of hand-designed cases can improve agent performance. In our experiments, fine-tuning a pre-trained model for just a few minutes enables agents to achieve near-perfect goal-reaching rates on previously difficult tasks (Section \ref{sec:finetuning}).  
\end{itemize}
As our agents may be independently interesting to use as part of other simulators or in autonomous vehicle test cases, we open-source our agents at \url{www.github.com/Emerge-Lab/gpudrive}.

We note three limitations of our work that provide some nuance. 
First, our benchmark, built atop the Waymo Open Motion Dataset, consists of short-horizon scenarios that are only 9 seconds long. Second, we excluded pedestrians, cyclists, and traffic lights. Finally, we train agents to optimize specific criteria above realism, we find that there is still ample room for improvement in making their behavior human-like, as indicated by the WOSAC meta-score of $~0.63$ (Section \ref{sec:realism_wosac}). An interesting direction for future work is to balance reliability with realism, ensuring agents meet performance standards while accurately reflecting human driving behavior across diverse scenarios.

While not explored in this paper, we anticipate that our findings extend to other domains such as neuroscience, where agent-based modeling is gaining momentum \citep{aldarondo2024virtual, johnson2024understanding, castro2025discovering}. Researchers are increasingly using physics-based simulators to create digital twins of animals, enabling cost-effective and controlled experimentation. For these agents to be useful models of animal behavior, reliability and robustness appear essential. For example, a rodent foraging model, should not exhibit free movement. We hope our work contributes to the improvement of agent-based modeling, helping to enhance controllability and robustness across different domains.




\bibliography{paper}
\bibliographystyle{plainnat}

\newpage
\appendix
\onecolumn





\section{Observation features and design choices}
\label{appendix:feature_details}

The observation at time step $t$ for agent $i$, $\mathbf{o}_i^t$, is multi-modal and consists of three types of information: the ego state, the visible view of the scene, and the partner observation. We set the maximum number of agents per scenario throughout the experiments, $N=64$. We limit agents to vehicles. A given agent's observation is provided as a flattened vector of $\sim 3000$ elements.

\begin{table}[htbp]
    \centering
    \caption{Ego state features and dimensions provided in the observation $o_t^i$.}
    \begin{tabular}{@{}p{0.2\linewidth}p{0.1\linewidth}p{0.65\linewidth}@{}} 
        \toprule
        Feature & Dimension & Description \\ \midrule
        Speed & 1 & The speed of the agent \\
        Vehicle length & 1 & Length of the agents' bounding box \\
        Vehicle width & 1 & Width of the agents' bounding box \\
        Relative goal position & 2 & Distance from agent to the target position in the $x$ and $y$ axis \\
        Collision state & 1 & Whether the agent is in collision (1) or not (0) \\ 
        \bottomrule
    \end{tabular}
    \label{tab:ego_feature_dimensions}
\end{table}

\begin{table}[htbp]
    \centering
    \caption{Visible view or road graph features and dimensions provided in the observation $o_t^i$. The road graph consists of a sampled set of $R$ nearest road points, where $R$ is set to 200 in the experiments.}
    \begin{tabular}{@{}p{0.2\linewidth}p{0.1\linewidth}p{0.65\linewidth}@{}} 
        \toprule
        Feature & Dimension & Description \\ \midrule
        $x$ & $1 \cdot R$  & Relative x coordinate of the road point  \\
        $y$ & $1 \cdot R$ & Relative y coordinate of the road point \\
        Segment length & $1 \cdot R$ & Length of the road segment associated with the ($x, y$) coordinate \\
        Segment width & $1 \cdot R$ & Width of the road segment associated with the ($x, y$) coordinate \\
        Segment height & $1 \cdot R$ & Height of the road segment associated with the ($x, y$) coordinate \\
        Segment orientation & $1 \cdot R$ & Angle between the segment midpoint and the ego agent  \\
        Type & $1 \cdot R$ & Integer indicating the type of the road point. Existing types are: Road edge (impassable; boundary of the road), road lane, road line, stop sign, crosswalk, and speed bump. Integers are one-hot encoded during training, which multiplies the feature dimension by the total number of classes. \\
        \bottomrule
    \end{tabular}
    \label{tab:rg_feature_dimensions}
\end{table}

\begin{table}[htbp]
    \centering
    \caption{Partner (``the other``) agent features and dimensions provided in the observation $o_t^i$. Partner information is visible within the observation radius.}
    \begin{tabular}{@{}p{0.2\linewidth}p{0.1\linewidth}p{0.65\linewidth}@{}} 
        \toprule
        Feature & Dimension & Description \\ \midrule
        Speed & $1 \cdot N - 1$ & The speed of the other agents \\
        ($\mathbf{x}, \mathbf{y}$) & $2 \cdot N-1$ & Relative positions of the other $N-1$ agents in the scene. Information is only provided if the partner agents are within the observation radius of the ego agent, and are left as zero otherwise.  \\
        ($\mathbf{\theta}_x, \mathbf{\theta}_y$) & $2 \cdot N-1$ & Relative orientation of the other $N-1$ agents in the scene. Information is only provided if the partner agents are within the observation radius of the ego agent, and are left as zero otherwise.  \\
        $(w, l, h)$ & $3 \cdot N - 1$ & The width, length, and height of the bounding boxes of the other agents. \\
        \bottomrule
    \end{tabular}
    \label{tab:partner_feature_dimensions}
\end{table}

\newpage
\section{Additional figures}

\subsection{Sample rollouts}

\begin{figure}[htbp]
    \centering
    \begin{subfigure}[t]{0.32\linewidth}
        \centering
        \includegraphics[width=\linewidth]{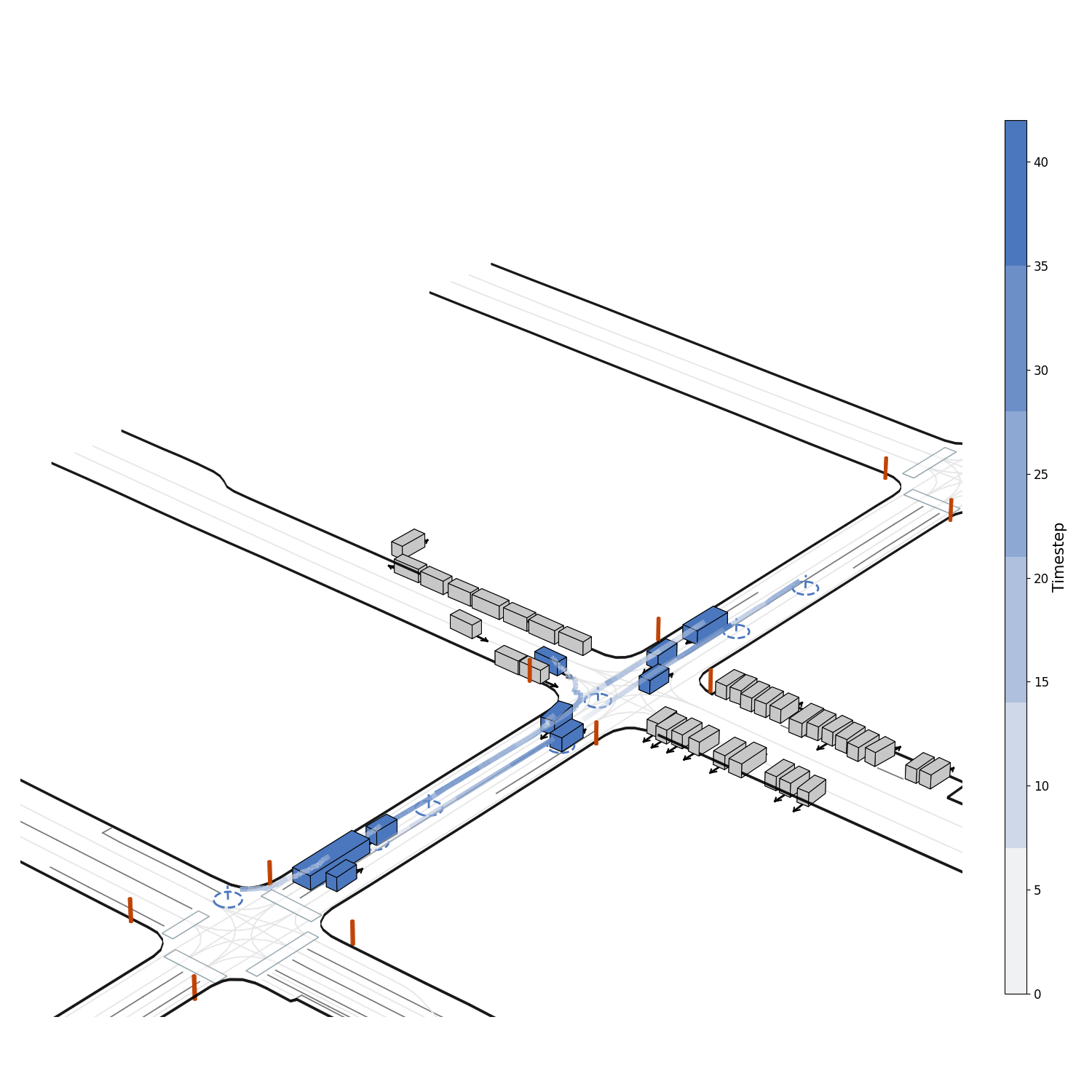}
        \caption{}
        \label{fig:sample_rollout_example_1}
    \end{subfigure}
    \hfill
    \begin{subfigure}[t]{0.32\linewidth}
        \centering
        \includegraphics[width=\linewidth]{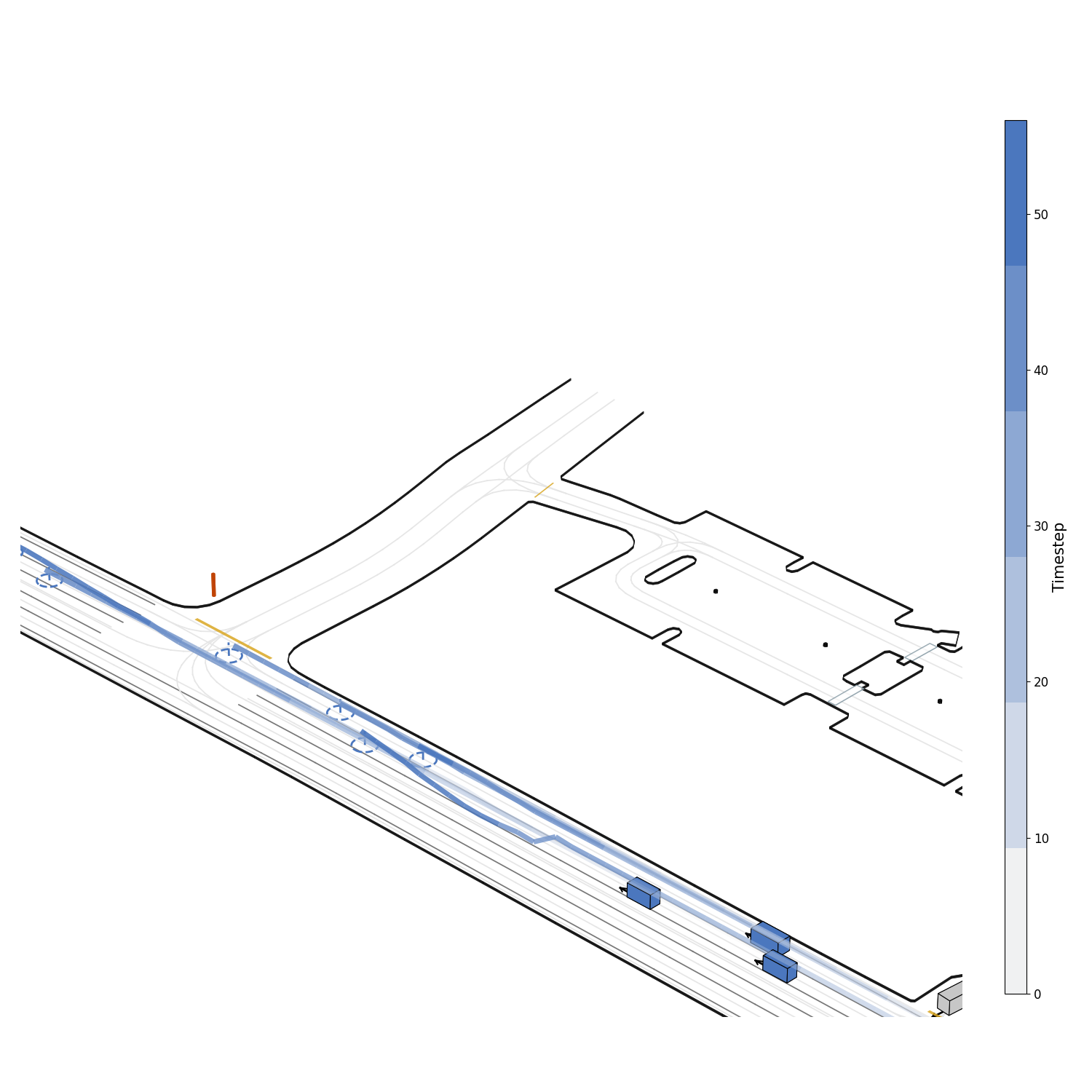}
        \caption{}
        \label{fig:sample_rollout_example_2}
    \end{subfigure}
    \hfill
    \begin{subfigure}[t]{0.32\linewidth}
        \centering
        \includegraphics[width=\linewidth]{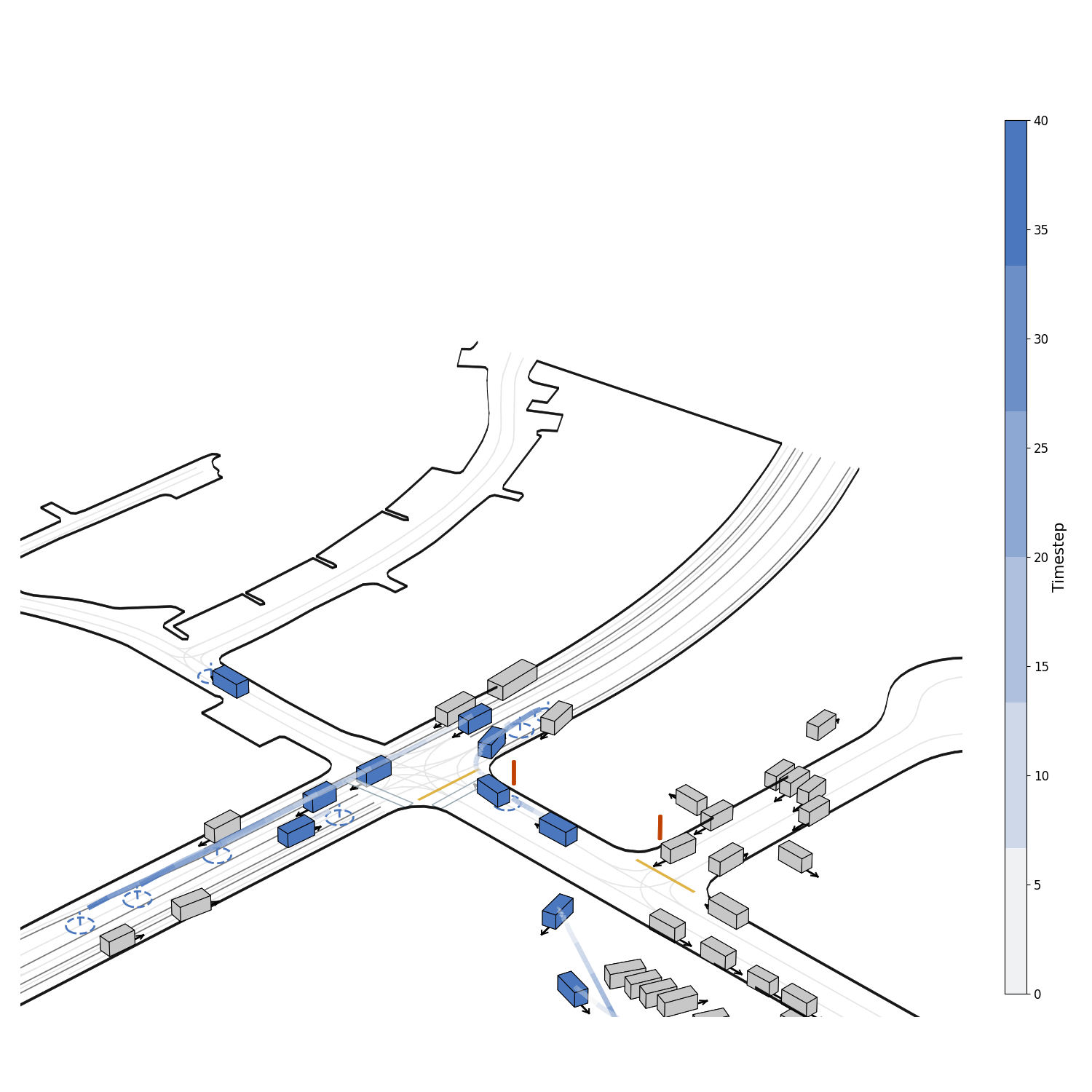}
        \caption{}
        \label{fig:sample_rollout_example_3}
    \end{subfigure}
\caption{Example rollouts with the best-trained policy. Agents controlled by the trained policy are shown in blue, while static agents are colored in grey.}
\label{fig:3d_examples}
\end{figure}

\begin{figure}[htbp]
    \centering
    \begin{subfigure}[t]{0.32\linewidth}
        \centering
        \includegraphics[width=\linewidth]{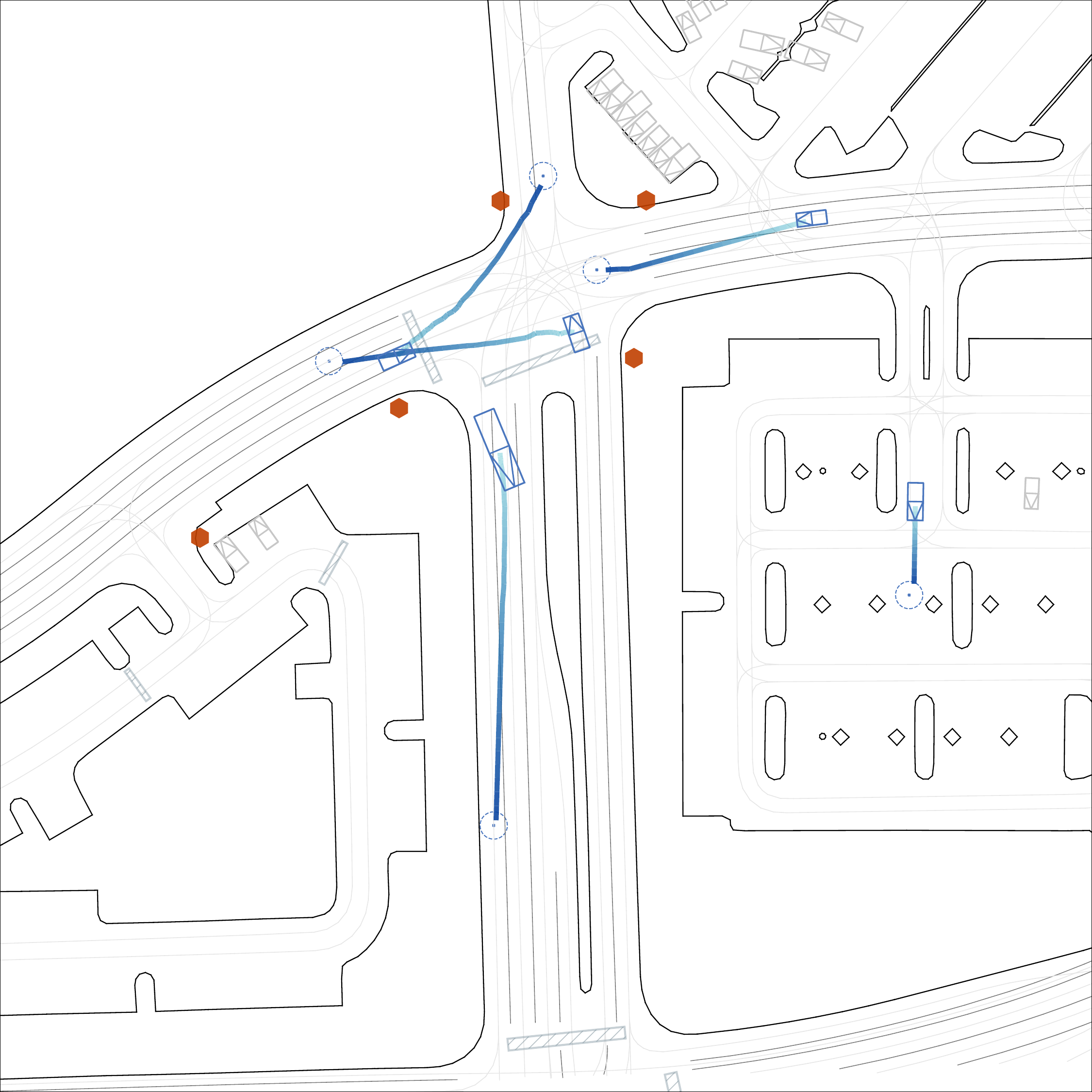}
        \caption{}
        \label{fig:example_1}
    \end{subfigure}
    \hfill
    \begin{subfigure}[t]{0.32\linewidth}
        \centering
        \includegraphics[width=\linewidth]{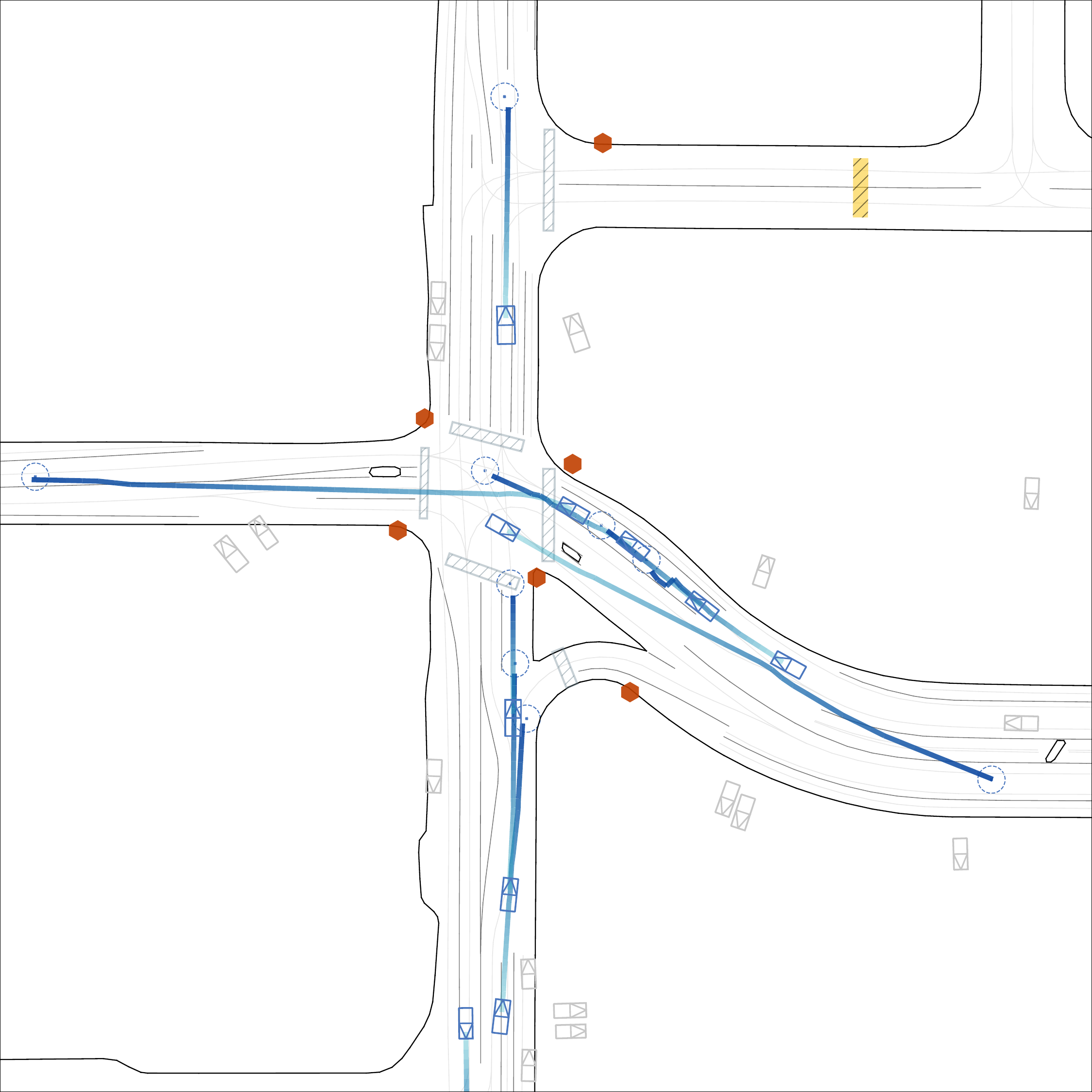}
        \caption{}
        \label{fig:example_2}
    \end{subfigure}
    \hfill
    \begin{subfigure}[t]{0.32\linewidth}
        \centering
        \includegraphics[width=\linewidth]{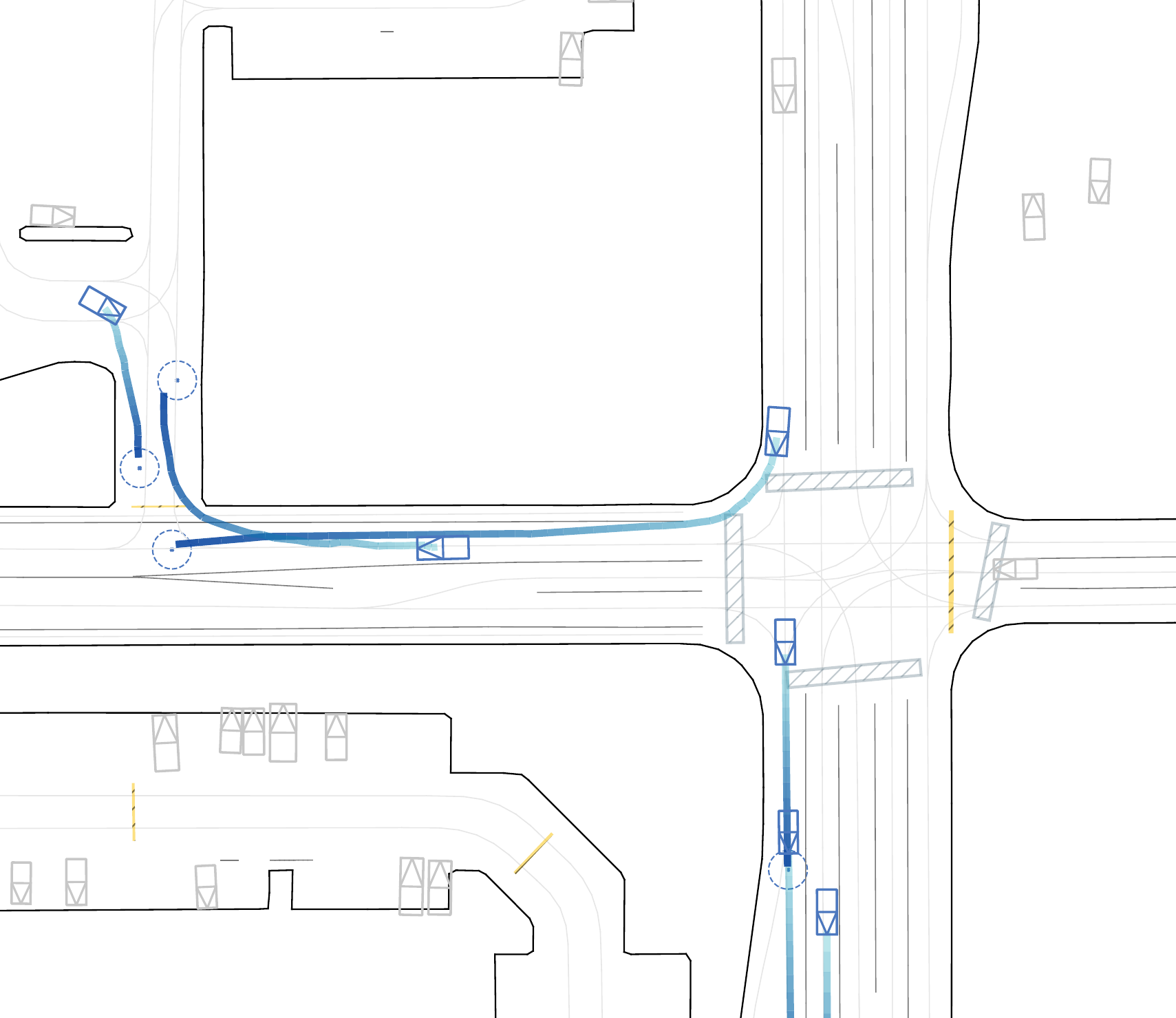}
        \caption{}
        \label{fig:example_3}
    \end{subfigure}
\caption{Example rollouts with the best-trained policy. Agents controlled by the trained policy are shown in blue, while static agents are colored in grey.}
\label{fig:2d_examples}
\end{figure}

\subsection{Neural Network Architecture}
\label{sec:neural_net}
\begin{figure}[htb]  
    \centering
    \includegraphics[width=0.7\linewidth]{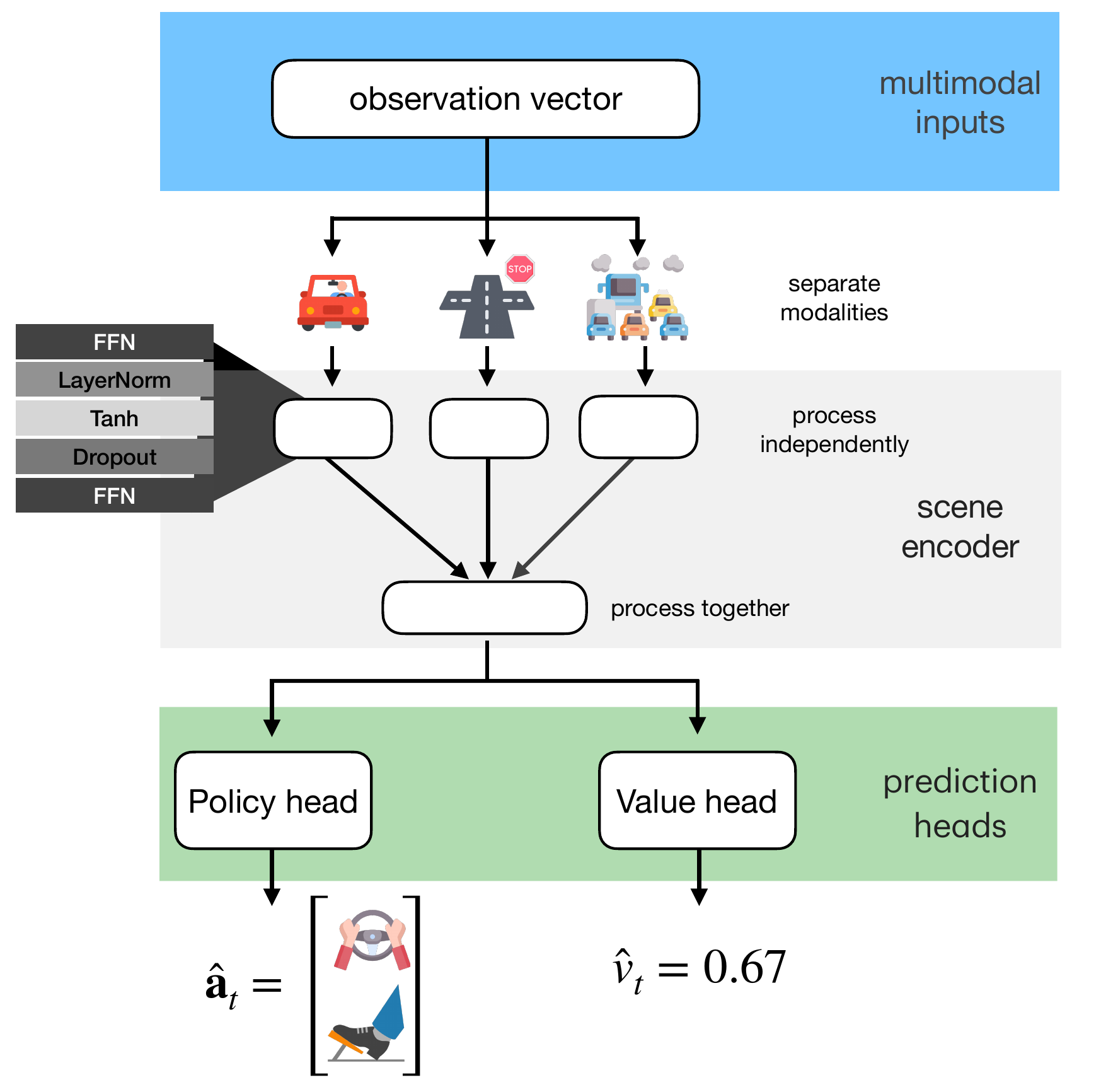} 
    \caption{\textbf{Network architecture}. The relative observation vector $o_t^i$ is first decomposed into its separate modalities: the ego state (i.e. the agent's information about itself and its goals), the visible portion of the road graph, and the speeds, yaws, and relative positions of the other agents in the scene. These modalities are first processed separately. Their outputs are combined and max pooled, then processed together. The hidden layer is finally fed into an actor and a critic head.}
    \label{fig:network_architecture}
\end{figure}

\section{Distribution of errors and remaining failure modes}
\label{sec:failure_modes}
\begin{figure}[htb]
    \centering
    \includegraphics[width=\linewidth]{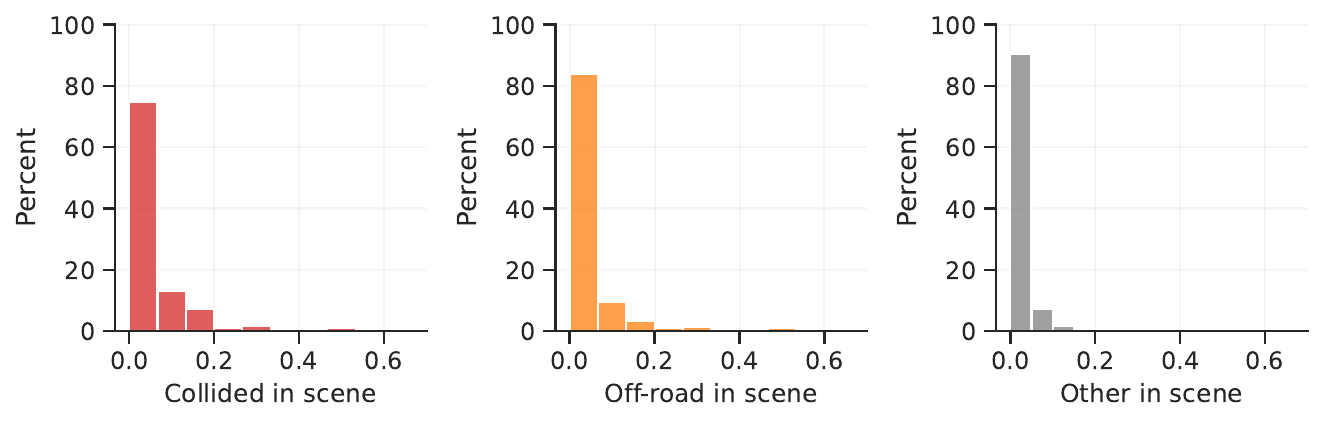}
    \caption{\textbf{Probability distribution function for each type of error for scenes that are not fully solved.}. \textit{Left}: Percentage of agents that collided. \textit{Middle}: Percentage of agents that went off-road. \textit{Right}: Percentage of agents that neither failed nor reached their goal. Note that almost all scenes contain just a single failure.}
    \label{fig:error_distributions}
\end{figure}

We analyze scenarios that are not perfectly solved, defined as those with a collision rate or off-road rate greater than 0, or where at least one agent fails to reach its goal. A selection of failure modes can be viewed on the \href{https://sites.google.com/view/reliable-sim-agents/home}{project page}. Together, these account for $8.95$\% of the test dataset ($896$ out of $10,000$ scenarios). Figure \ref{fig:error_distributions} shows the histogram of error distributions, revealing that most unsolved scenarios have only a small error rate. We compute the Pearson correlation between off-road fractions and collision rates to examine potential relationships between failure modes. The result, $\rho = 0.0135$, is not significant at $\alpha = 0.05$, indicating no meaningful correlation between these two metrics in the unsolved scenarios and suggesting that errors are spread across scenarios.

Additionally, we analyze the top $0.5$\% failure modes in each category (collision rates, off-road rates, and agents that did not reach the goal position) of the test set. This analysis provides information about challenging aspects of these scenarios. The key takeaways are as follows.

\textbf{Rare map layouts and objects:} High off-road rates occur in scenarios with rarely occurring road structures. One example of this is roundabouts. A large fraction ($15$\%) of the top fraction of collision rates was in roundabout scenes. The rest included road layouts that are simply harder to navigate, such as tight corners, narrow lane entries, parking lots, etc. Larger vehicles especially struggle with such maps. This coupled with multiple vehicles crowding leads to some of them going off-road.

\textbf{Coordination:} High collision rates occur in intersections, speedy highways, and crowded scenes where sophisticated interaction is required (eg: letting another agent pass before you, making space for another agent to overtake, etc). Crowding and interaction coupled with rare map layouts compound the difficulty of the scene and lead to a higher collision and off-road rate.

\textbf{Out of time:} Some agents have goals further away than others. Having a finite horizon of $91$ steps means trying to squeeze past agents and narrow lanes when it is very hard to. This leads to a higher collision and off-road rate compared to scenes with closer goals. This can also compound difficulty in scenes with the aforementioned properties. 

\section{Considerations for learning sim agents through self-play PPO}

\subsection{Collision behavior}
\label{appendix:collision_behavior}

GPUDrive supports three types of collision behaviors: \textit{ignore}, \textit{remove}, and \textit{stop}. Each of these has different effects on the types of behaviors agents learn over time. We briefly outline some things to be aware of below, which might be useful for future experiments. 

\paragraph{Ignoring collisions} When collision behavior is ignored, the agent is not terminated when it collides with another agent or touches a road edge. As such, it can proceed to the goal and collide within a single episode. To discourage collisions, it seems reasonable to give agents a penalty. However, since, in most scenarios, the probability of getting negative signals in an episode with random behavior (e.g. hitting a road edge) is significantly larger than the probability of receiving a positive signal (getting to the goal), the value function may become overly pessimistic because the majority of the advantages the agent is receiving will be negative, and as such the probability of actions that lead to these negative advantages, such as higher acceleration, will be decreased. This can lead to a behavior where agents freeze (they learn to stay on the road) and do not head towards the goal. This can be avoided by ensuring that agents receive enough positive signals along with negative ones, especially early on during learning. This can be achieved by sufficient exploration through a large enough entropy coefficient.

\paragraph{Removing agents at collision} Another option is to simply terminate agents whenever they do something that is not desired (in our case colliding) without assigning penalties (giving negative rewards). This means that the goal can only be achieved if the agent does not do something bad. Since the penalty in this case is implicit, the value function can not become overly pessimistic. Instead, the advantages will be 0 most of the time early on in training. Once the first positive signals are achieved by accident (which is inevitable given the small maps of the WOMD and a high enough entropy coefficient), the probability of the right action sequences will be increased until all agents hit their goals without colliding or going off-road.

\begin{table}[htbp]
    \centering
    \caption{Overview of collision behaviors}
    \begin{tabular}{@{}p{0.15\linewidth}p{0.4\linewidth}p{0.4\linewidth}@{}} 
        \toprule
        Collision behavior & Pro's & Caveats \\ \midrule
        Ignore & 1) Agents receive a diverse range of observations & 1) Value function can become overly pessimistic; 2) Large exploration space: A large set of possible states leads to agents seeing lots of useless observations during exploration \\
        Stop & 1) Closest to the real-world effect of collisions & 1) Introduces extra challenge during learning: drive around other stopped agents \\
        Remove & 1) Simple & 1) Might lead to unrealistic behavior when used and agents are not removed from the scene; 2) Might be difficult to reach certain states because the agents are always removed upon collisions; 3) Number of completed episodes is large in the beginning since most agents are terminated within 10 steps and subsequently a lot of \texttt{reset()} calls early on in training, which decreases the SPS. \\
        \bottomrule
    \end{tabular}
    \label{tab:collision_behaviors}
\end{table}

\section{Analyses.}
\label{appendix:ood}

\subsection{Detecting out-of-distribution events}
To find rare goals, we analyzed the full training dataset ($\approx 129,000$ scenes) or about $4.2$ million controllable agents. Of these, we found approximately $30,000$ agents ($0.73$\%) making a U-turn, and $47,000$ agents ($1.13$\%) driving in reverse (see Appendix~\ref{appendix:ood} for the exact definition of these events). Further, most agents driving in reverse were simply pulling out of park, with goals immediately behind them. We observed a distinct lack of goals where the agent needs to execute a complex U-turn, making it plausibly out of distribution. We then hand-designed 13 scenarios from the test dataset with a total of 27 agents across all scenes, placing goals behind agents. This was done by setting the new goal for each agent to $(x_f - x_i, y_f - y_i)$, where $(x_i, y_i)$ is the initial position, and $(x_f, y_f)$ is the original goal. We chose the scenes in such a way that doing this process for all controlled agents results in valid and reachable goals. Figure \ref{fig:finetuning_steps}.
\begin{enumerate}
    \item \textbf{U-turn}: For each time step $t$ where the agent is valid, we check the condition: abs(heading[t] - heading[initial]) > 150\degree.
    \item \textbf{Driving in reverse}: For each time step $t$ where the agent is valid, calculate the direction of its velocity vector and subtract it from its heading angle. If the absolute difference is greater than a threshold (150\degree), it is driving in reverse. Note: We only detect driving in reverse if it occurs for more than a threshold (10) consecutive steps, and above a minimum magnitude velocity (0.5 km/hr). 
\end{enumerate}

\subsection{Placing goals behind agents}

\begin{table}[htbp]
\centering
\caption{Aggregate performance comparison between Altered and Original goal positions (mean $\pm$ std).}
\small 
\begin{tabular}{@{}lcccc@{}} 
\toprule
\textbf{Class} & \textbf{Goal achieved} \textcolor{RoyalBlue}{\textuparrow} & \textbf{Collided} \textcolor{BrickRed}{\textdownarrow} & \textbf{Off-road} \textcolor{RedOrange}{\textdownarrow} & \textbf{Other} \textcolor{lightgray}{\textdownarrow} \\
\midrule
Altered  & 53.5 $\pm$ 38.4 & 10.0 $\pm$ 8.3 & 6.7 $\pm$ 16.1 & 41.7 $\pm$ 32.0 \\
Original & 100.0 $\pm$ 0.0 & 0.0 $\pm$ 0.0 & 0.0 $\pm$ 0.0 & 0.0 $\pm$ 0.0 \\
\bottomrule
\end{tabular}%
\label{tab:goals_behind_agents_perf}
\end{table}





\section{PPO implementation details.}

\subsection{Hyperparameters}

Table \ref{tab:PPO_hparams} reports the hyperparameters used for the results in our experiments.

\begin{table}[htbp]
    \centering
    \caption{PPO Algorithm Hyperparameters}
    \begin{tabular}{@{}p{0.25\linewidth}p{0.13\linewidth}p{0.52\linewidth}@{}} 
        \toprule
        Parameter & Value & Description \\ \midrule
        \texttt{total\_timesteps} & 1,000,000,000 & Total number of timesteps for training. \\
        \texttt{batch\_size} & 524,288 & Number of timesteps collected in each rollout. \\
        \texttt{minibatch\_size} & 16,384 & Number of timesteps in each minibatch for gradient updates. \\
        \texttt{learning\_rate} & 3e-4 & Initial learning rate for the optimizer. \\
        \texttt{anneal\_lr} & \texttt{false} & Whether to anneal the learning rate over time. \\
        \texttt{gamma} & 0.99 & Discount factor for future rewards. \\
        \texttt{gae\_lambda} & 0.95 & Lambda parameter for Generalized Advantage Estimation. \\
        \texttt{update\_epochs} & 2 & Number of epochs to update the policy network per rollout. \\
        \texttt{norm\_adv} & \texttt{true} & Whether to normalize advantages during training. \\
        \texttt{clip\_coef} & 0.2 & PPO clipping coefficient for policy updates. \\
        \texttt{clip\_vloss} & \texttt{false} & Whether to clip the value loss. \\
        \texttt{vf\_clip\_coef} & 0.2 & Clipping coefficient for value function updates. \\
        \texttt{ent\_coef} & 0.0001 & Entropy regularization coefficient to encourage exploration. \\
        \texttt{vf\_coef} & 0.5 & Coefficient for the value function loss in the total loss. \\
        \texttt{max\_grad\_norm} & 0.5 & Maximum norm for gradient clipping. \\
        \texttt{target\_kl} & \texttt{null} & Target KL divergence for policy updates (unused if null). \\
        \texttt{collision\_weight} & -0.75 & Penalty weight for collision events. \\
        \texttt{off\_road\_weight} & -0.75 & Penalty weight for off-road events. \\
        \texttt{goal\_achieved\_weight} & 1.0 & Reward weight for achieving the goal. \\
        \bottomrule
    \end{tabular}
    \label{tab:PPO_hparams}
\end{table}

\section{Compute resources}

Experiments were run on either a single NVIDIA A100 or an RTX4080 device for 12-36 hours per experiment. Including hyperparameter tuning and experimentation, all runs combined for this paper took approximately 5 GPU days.

\end{document}